\documentclass[british]{article}

\usepackage{spconf}
\usepackage{times}
\usepackage{epsfig}
\usepackage{graphicx}
\usepackage{amsmath}
\usepackage{amssymb}
\usepackage{lipsum}  
\usepackage[comma,sort&compress,numbers]{natbib}
\usepackage[small]{caption}
\usepackage{makecell}
\usepackage{multirow}
\usepackage{booktabs,soul}
\usepackage{tabularx}
\usepackage[dvipsnames]{xcolor}
\usepackage{enumitem}
\usepackage{subfig}
\usepackage{float}
\usepackage{array}

\newcommand\norm[1]{\left\lVert#1\right\rVert}
\newcolumntype{P}[1]{>{\centering\arraybackslash}p{#1}}
\newcolumntype{M}[1]{>{\centering\arraybackslash}m{#1}}
\usepackage[breaklinks,colorlinks]{hyperref}
\usepackage{tabularx}
\usepackage{flushend}

\begin{document}

%%%%%%%%% TITLE
\title{Enhancing Deformable Convolution based Video Frame Interpolation with Coarse-to-fine 3D CNN}
\name{Duolikun Danier, Fan Zhang and David Bull \thanks{The authors acknowledge the fundings from China Scholarship Council, the University of Bristol and the
MyWorld Strength in Places Programme.}}
\address{Visual Information Laboratory, University of Bristol, One Cathedral Square, BS1 5DD, United Kingdom\\
% Bristol, BS8 1UB, United Kingdom \\ 
\{Duolikun.Danier, Fan.Zhang, Dave.Bull\}@bristol.ac.uk}

\maketitle
%\thispagestyle{empty}

%%%%%%%%% ABSTRACT
\begin{abstract}
   This paper presents a new deformable convolution-based video frame interpolation (VFI) method, using a coarse to fine 3D CNN to enhance the multi-flow prediction. This model first extracts spatio-temporal features at multiple scales using a 3D CNN, and estimates multi-flows using these features in a coarse-to-fine manner. The estimated multi-flows are then used to warp the original input frames as well as context maps, and the warped results are fused by a synthesis network to produce the final output. This VFI approach has been fully evaluated against 12 state-of-the-art VFI methods on three commonly used test databases. The results evidently show the effectiveness of the proposed method, which offers superior interpolation performance over other state of the art algorithms, with PSNR gains up to 0.19dB. 
\end{abstract}


%%%%%%%%% BODY TEXT
\section{Introduction}

In video processing, video frame interpolation (VFI) is both an important tool and a popular research topic. VFI refers to the task of generating one or more intermediate frames between every two original frames in a video sequence. It is used in a wide range of applications, including frame rate up-conversion~\cite{bao2018high}, slow-motion generation~\cite{jiang2018super}, novel view synthesis~\cite{flynn2016deepstereo} and video compression~\cite{wu2018video}. Interpolating high-quality frames requires accurate motion modelling and preservation of spatiotemporal consistencies, which becomes more difficult in scenarios that involve large and complex motions.

Existing VFI methods are usually subdivided into two main categories: flow-based~\cite{liu2017video, jiang2018super, bao2019depth, park2020bmbc, sim2021xvfi} and kernel-based~\cite{lee2020adacof, gui2020featureflow,  niklaus2017video, ding2021cdfi, kalluri2020flavr}. Flow-based methods typically use optical flows to warp input frames. Although such one-to-one pixel mapping based approaches can produce sharp results and work well on less complex scenes, they have limitations when handling more complex motions. On the other hand, kernel-based methods learn convolution kernels to synthesise each output pixel from multiple input pixels, corresponding to a many-to-one mapping. Among these kernel-based methods, deformable convolution (DefConv)~\cite{dai2017deformable} has been adopted more and more frequently~\cite{lee2020adacof, gui2020featureflow, cheng2021multiple, ding2021cdfi} due to its flexibility in motion modelling. DefConv-based VFI methods estimate a local kernel and multiple flow vectors for each output pixel. The output pixel is then synthesised by convolving the predicted kernel with the input pixels pointed by these flow vectors -- these local kernels and flow vectors are denoted as \textit{multi-flows} in the literature~\cite{gui2020featureflow}. Compared to one-to-one flows, this type of many-to-one mapping enables more complex transformations, hence improving the ability to model complex motions.

It is noted that most of the existing DefConv-based methods use only two adjacent frames to interpolate the middle frame, whereas the use of a larger window can provide richer information on the spatio-temporal characteristics of the input signal~\cite{kalluri2020flavr,danier2021spatio}. Moreover, when predicting multi-flows under the DefConv-based VFI framework, only 2D convolutional neural networks (CNNs) have been employed, which cannot provide an explicit way of extracting temporal information from the content. Performing spatio-temporal filtering using 3D CNNs, however, has been shown for other applications~\cite{yang2021spatiotemporal} to be more effective in capturing content dynamics.

%=====================================================================
\begin{figure*}[t]
\begin{center}
   \includegraphics[width=0.95\linewidth]{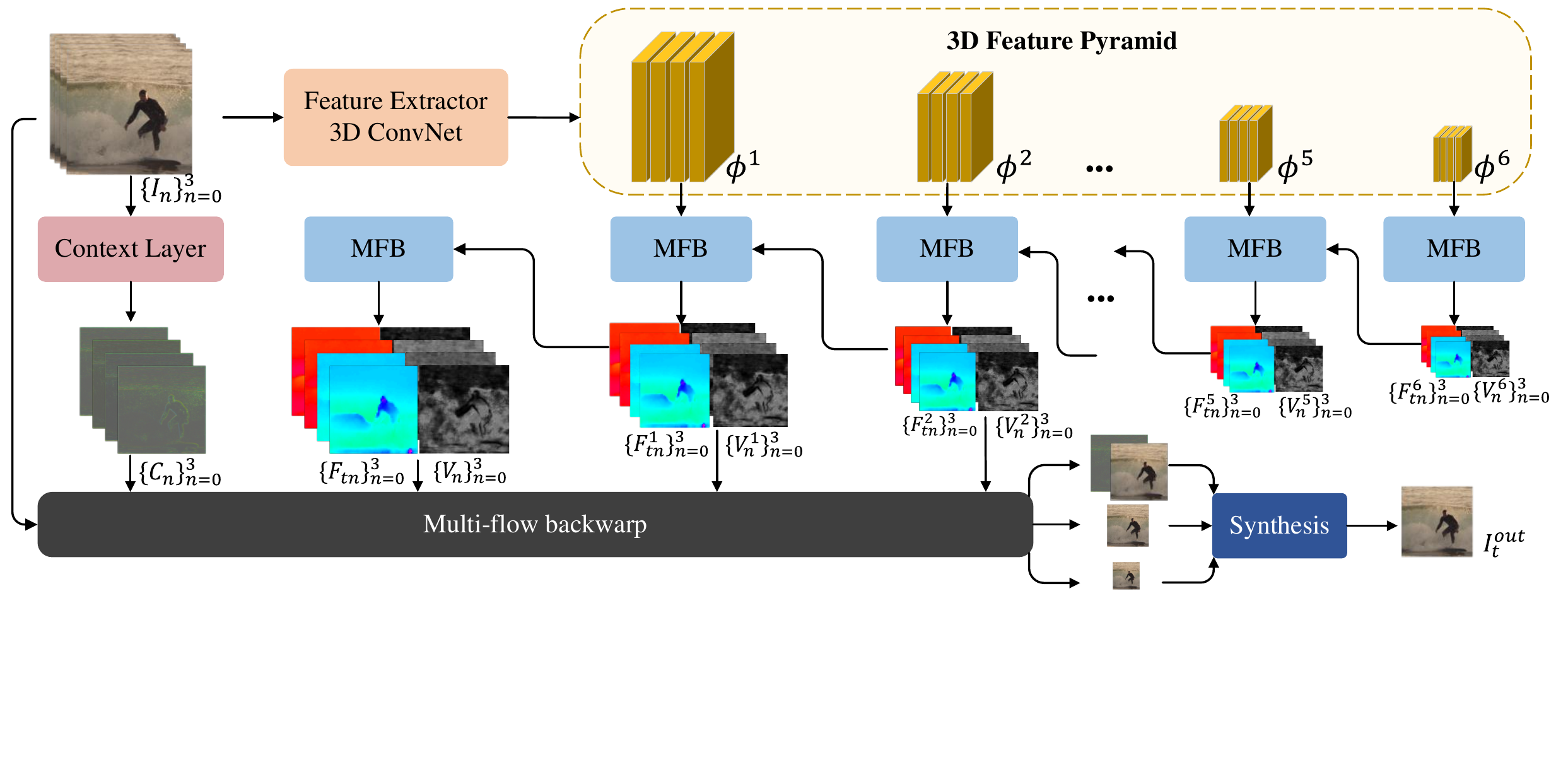}
\end{center}
\vspace{-25mm}
\caption{\label{fig:overall}
Overall architecture of the proposed model.}
\vspace{-2mm}
\end{figure*}
%=====================================================================
%=====================================================================
\begin{figure}[t]
\begin{center}
   \includegraphics[width=0.9\linewidth]{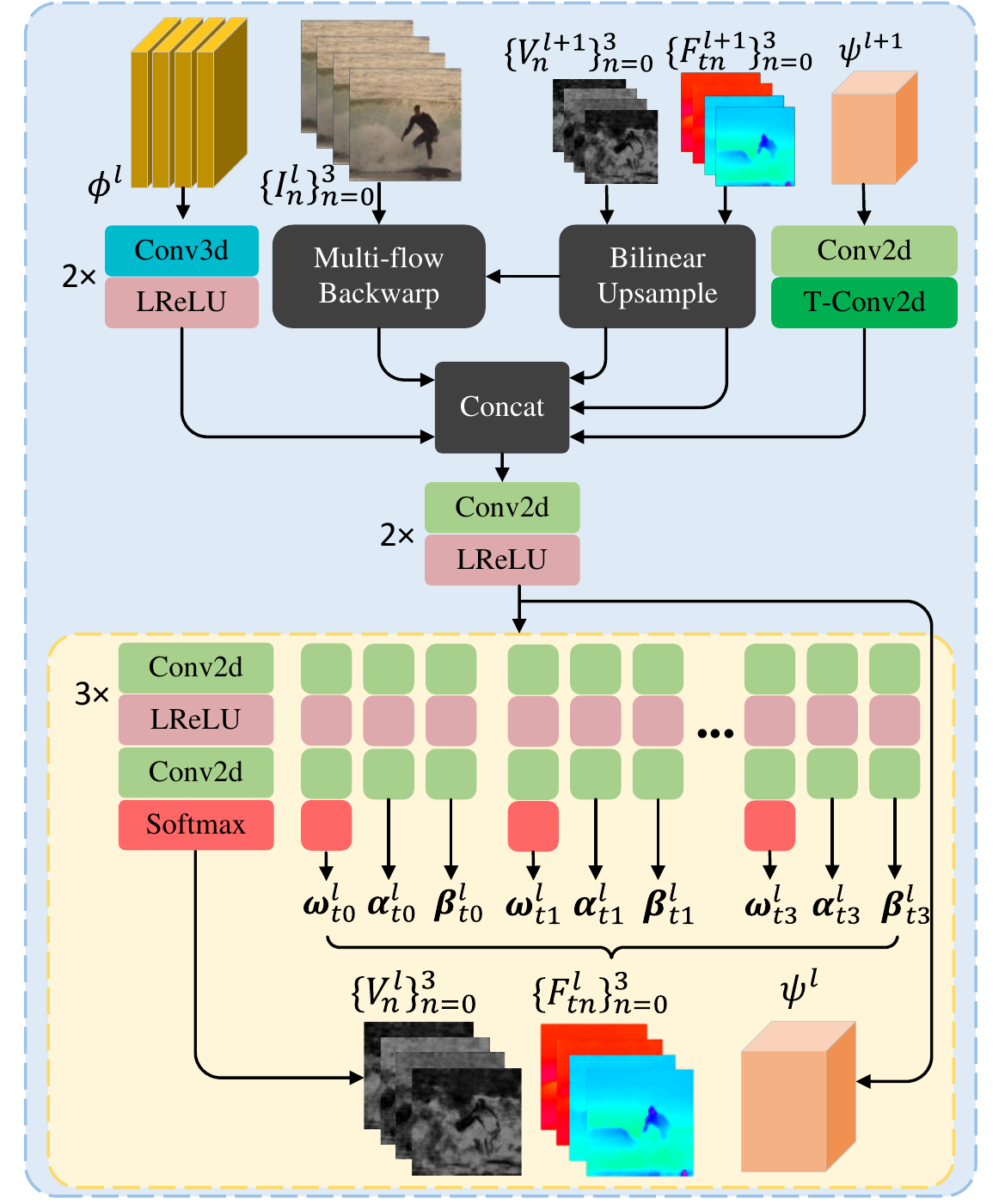}
\end{center}
\vspace{-4.5mm}
\caption{\label{fig:detail}
The architecture of the multi-flow block (MFB).}
\vspace{-4mm}
\end{figure}
%=====================================================================

In this context, this paper presents a novel coarse-to-fine 3D CNN architecture for DefConv-based video frame interpolation. The proposed model first predicts multi-flows for DefConv-based warping at a low (coarse) resolution scale and refines them over multiple scales. In contrast to previous works that use such coarse-to-fine scheme~\cite{sun2018pwc,sim2021xvfi}, our network involves 3D convolutions that perform filtering in both spatial and temporal domains to better capture content dynamics. The effectiveness of the proposed network has been demonstrated through ablation studies and  comprehensive quantitative experiments. The results show that our proposed model delivers state-of-the-art performance on various test datasets. 

The rest of this paper is organised as follows. The proposed model is described in detail in Section~\ref{sec:method}. Section~\ref{sec:exp} first specifies the setup of the ablation and benchmark experiments, and then presents the experimental results alongside analysis. Finally, Section~\ref{sec:conclusion} concludes the paper.

\section{Proposed Method}\label{sec:method}
The overall architecture of the proposed network is shown in Fig.\ref{fig:overall}. The network takes the original frames $I_0, I_1, I_2, I_3$ in a video as input, and outputs $I^\mathrm{out}_t$ where $t=1.5$. Given the four input frames, our model predicts the DefConv-based multi-flows from non-existent $I^\mathrm{out}_t$ to all the four input frame using a coarse-to-fine 3D CNN. For each input frame, a context map is also extracted, which has been previously reported to improve overall VFI performance~\cite{niklaus2018context, bao2019depth, park2020bmbc}. The input frames and context maps are then warped using the predicted multi-flows, and passed into the synthesis network to produce the interpolated result.

% \noindent\textbf{DefConv-based multi-flows.} 
\subsection{DefConv-based multi-flows}

Here we briefly introduce the multi-flows in the DefConv-based VFI framework~\cite{lee2020adacof}, which is the basis of our method. The multi-flow from the desired output frame $I_t$ (of size $H\times W$) to one of the input frames $I_n$, where $n=0,1,2,3$, is defined as  $F_{tn} = (\boldsymbol{\alpha}, \boldsymbol{\beta}, \boldsymbol{\omega})$. Here $\boldsymbol{\alpha}, \boldsymbol{\beta} \in \mathbb{R}^{H\times W\times M}$ are the x-, y-components of $M$ flow vectors from $I_t$ to $I_n$, and $\boldsymbol{\omega}\in [0,1]^{H\times W\times M}$ specifies the kernel of size $M$ at each spatial location ($\sum_{k=1}^M \boldsymbol{\omega}(x,y,k)=1$). That is, for each pixel location $(x,y)$ in $I_t$, $F_{tn}$ specifies $M$ flow vectors and an $M$-sized weighting kernel. These parameters can be used to produce a warped frame $\hat{I}_t$ from $I_n$ with:

\small
\begin{equation}
\label{eqn:multiflow}
    \hat{I}_{tn}(x,y) = \sum_{k=1}^M \boldsymbol{\omega}(x,y,k)I_n(x+\boldsymbol{\alpha}(x,y,k), y+\boldsymbol{\beta}(x,y,k))
\end{equation}

\normalsize
Given $I_0, I_1, I_2, I_3$, the network predicts the multi-flows $\{F_{tn}\}_{n=0}^3$ from the intermediate frame $I_t$ to the inputs, so that the input frames can be warped to time $t$ using (\ref{eqn:multiflow}) to produce $\{\hat{I}_{tn}\}_{n=0}^3$. We extend the occlusion reasoning in~\cite{jiang2018super,lee2020adacof} to four frames and additionally predict visibility maps $\{V_n \in [0,1]^{H\times W}\}_{n=0}^3$, where $\sum_{n=0}^3 V_n(x,y)=1$, to combine the four warped candidates and produce a single frame $\hat{I}_t$:
\begin{equation}
\label{eqn:occlusion}
    \hat{I}_t = \sum_{n=0}^3 V_n \odot \hat{I}_{tn}
\end{equation}
where $\odot$ represents Hadamard multiplication. 

% \noindent\textbf{Network architecture.} 
\subsection{Network Architecture}

As shown in Fig.\ref{fig:overall}, firstly, the feature extractor performs spatiotemporal filtering to extract 3D features at $L$ scales indexed by $l$, where $l=0$ represents the original scale and $l=i$ corresponds to spatial down-sampling by $2^i$. The feature extractor employs the residual blocks in the ResNet3D~\cite{tran2018closer}. Specifically, each residual block consists of two 3D convolutional (Conv3d) layers with a residual connection, where the first Conv3d layer uses a spatial stride of 2 and all the remaining spatial and temporal strides are set to 1. Therefore, each residual block filters its input $\phi^{l-1}$ (for $l=1$ the input is the stacked input frames) in both spatial and temporal dimensions and outputs a feature map $\phi^l$ that is spatially down-sampled by a factor of 2. Compared to 2D convolutions that have been commonly used in previous methods, the spatio-temporal filtering additionally enables capturing of temporal characteristics of the frame sequence. A feature pyramid is then constructed using the output of each residual block, $\phi^l$, and passed to a series of Multi-Flow Blocks (MFBs).

% ===============================================================
\begin{table}[t]
\caption{Results (PSNR/SSIM) of ablation study.}
\resizebox{\linewidth}{!}{
\begin{tabular}{lcccccc}
\toprule
     & \multicolumn{2}{c}{UCF101} & \multicolumn{2}{c}{DAVIS}& \multicolumn{2}{c}{VFITex} \\ \cmidrule(lr){2-3}  \cmidrule(lr){4-5}  \cmidrule(lr){6-7}
 & PSNR & SSIM & PSNR & SSIM & PSNR & SSIM \\ \midrule
Ours-\textit{2D}   & 33.119 & 0.969  &26.510 & 0.848 & 27.223 & 0.886 \\  \midrule
$L_0$=0 & 33.305 & 0.970 & 27.383 & 0.873  & 28.346 & 0.914 \\ 
$L_0$=1 & 33.348 & 0.970 & 27.497 & 0.876  & 28.353 & 0.912 \\
$L_0$=2 & \textcolor{blue}{33.365} & \textcolor{blue}{0.970} & \textcolor{blue}{27.523} & \textcolor{blue}{0.877}  & \textcolor{blue}{28.459} & \textcolor{blue}{0.915} \\ \midrule
Ours & \textcolor{red}{33.431} & \textcolor{red}{0.970} & \textcolor{red}{27.633} & \textcolor{red}{0.878}  & \textcolor{red}{28.540} & \textcolor{red}{0.916} \\
\bottomrule
\end{tabular}
}
\vspace{-3mm}
\label{tab:ablation}
\end{table}
% ==============================================================

Starting from the scale $L_0 < L$, the MFBs predict a set of parameters $\{F_{tn}^l, V_n^l\}_{n=0}^3$ for multi-flow warping at each level in a coarse-to-fine manner, as employed in~\cite{sun2018pwc,sim2021xvfi}, to enhance the ability to capture large motions. The operations inside an MFB are shown in Fig.~\ref{fig:detail}. The input to an MFB at level $l$ consists of a 2D feature map $\psi^{l+1}$, the multi-flows $\{F_{tn}^{l+1}, V_n^{l+1}\}_{n=0}^3$ from the previous level, as well as the 3D feature map $\phi^l$ and the down-sampled original input frames $\{I_n^l\}_{n=0}^3$. The flow information from the previous level is up-sampled bilinearly and used to warp the down-sampled input frames using (\ref{eqn:multiflow}). The warped frames and the feature maps, combined with the up-sampled flow parameters, are then passed into a series of convolution layers to predict refined multi-flows for the current level.

\noindent\textbf{Context-awareness.} Previous works~\cite{niklaus2018context, bao2019depth} have shown the effectiveness of extracting contextual information for VFI. Therefore, we also incorporate a context layer to extract context maps $\{C_n\}_{n=0}^3$ from the input frames $\{I_n\}_{n=0}^3$. Specifically, we adopt the context layer proposed in~\cite{niklaus2018context} which is the first layer of a ResNet-18~\cite{he2016deep}.

\noindent\textbf{Synthesis network.} In the final synthesis step, the multi-flow parameters $\{F^l_{tn}, V^l_n\}_{n=0}^3$ at the three highest resolution scales, $l=0,1,2$, are used to warp the down-sampled input frames $\{I^{l}_{n}\}_{n=0}^3$ using (\ref{eqn:multiflow})-(\ref{eqn:occlusion}) to produce candidate frames $\{\hat{I}^{l}\}_{l=0}^{2}$. The context maps are also warped and denoted as $\hat{C}$. The synthesis network then uses these warped candidates to produce the interpolation result. Since it has been shown in \cite{niklaus2018context} that the GridNet~\cite{fourure2017residual} has superior ability to fuse multi-scale information, we adopt it as our synthesis network. 

% \noindent\textbf{Loss functions.} 
\subsection{Loss functions}

Our model is trained end-to-end by matching its output $I^\mathrm{out}_t$ with the ground-truth intermediate frame $I^\mathrm{gt}_t$ using the Laplacian pyramid loss~\cite{bojanowski2017optimizing}. The loss is defined as follows.
\begin{equation}
    \mathcal{L}_{lap} = \sum_{s=1}^{5} 2^{s-1} \norm{L^s(I^\mathrm{out}_t)-L^s(I^\mathrm{gt}_t)}_1
\end{equation}
where $L^s(\cdot)$ denotes the $s$\textsuperscript{th} level of the Laplacian pyramid. In addition, we match the warped frames $\hat{I}_t^l$ produced at lower scales with the ground truths $I_t^l$:
\begin{equation}
    \mathcal{L}_{charb} = \sum_{l=1}^{L_0} 2^{l} \Phi(\hat{I}_t^l-I_t^l)
\end{equation}
where $\Phi(x)=\sqrt{x^2+0.001^2}$ is the Charbonnier function. These two losses are combined to obtain the final loss function:
\begin{equation}
\label{eqn:loss}
    \mathcal{L} = \mathcal{L}_{lap} + \lambda \mathcal{L}_{charb}
\end{equation}
where $\lambda$ is a weighting hyper-parameter.

\section{Results and Analysis}\label{sec:exp}
In this section we analyse our proposed model through ablation study, and compare it with state-of-the-art VFI methods.

\subsection{Experimental Setup}\label{sec:expsetup}
\noindent\textbf{Implementation details.}
In our implementation, we set the coarsest level $L=6$, the starting level for predicting multi-flow $L_0=3$ and the number of flows $M=25$ (which is the default value of the original multi-flows in \cite{lee2020adacof}). The network is trained using $\mathcal{L}$ ($\lambda=0.01$), with the AdaMax optimizer~\cite{kingma2014adam} ($\beta_1=0.9,\beta_2=0.999$). The learning rate is initially 0.001 and reduced by a factor of 0.5 if the validation performance plateaus for 5 epochs. We perform instance normalisation~\cite{ulyanov2016instance} which has been found in \cite{niklaus2018context, kalluri2020flavr} to perform better than batch normalisation. The network is trained for 70 epochs using a batch size of 2. All training and evaluation were performed on a NVIDIA P100 GPU.

%=====================================================================
\begin{figure*}[t]
% \vspace{-10mm}
	\begin{center}
		\subfloat[Overlay]
		{\includegraphics[width=0.161\linewidth]{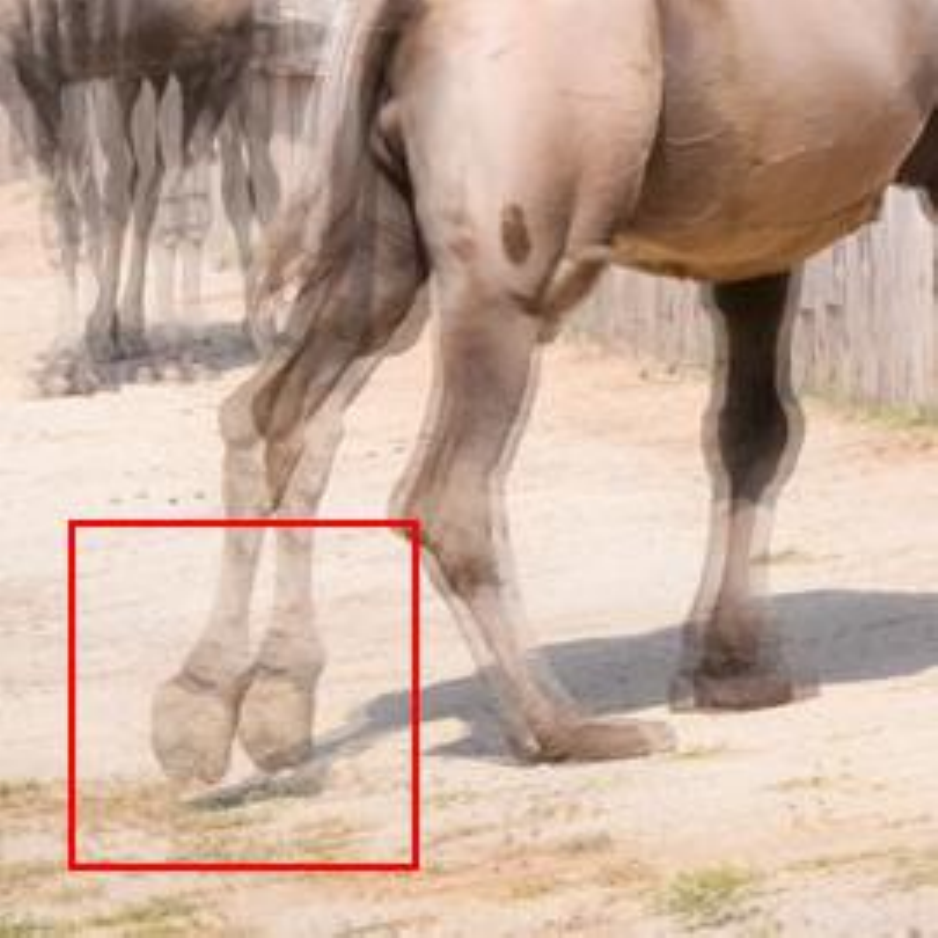}}\;\!\!
        \subfloat[EDSC]
    	{\includegraphics[width=0.161\linewidth]{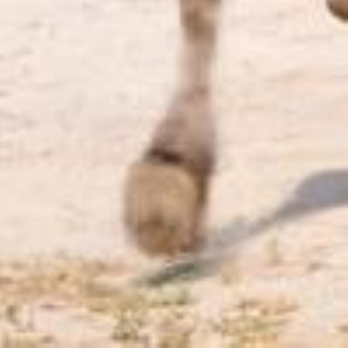}}\;\!\!
    	\subfloat[DAIN]
    	{\includegraphics[width=0.161\linewidth]{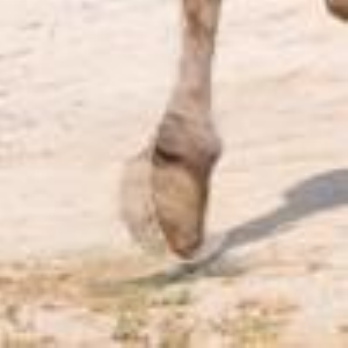}}\;\!\!
    	\vspace{-3mm}
    	\subfloat[FLAVR]
    	{\includegraphics[width=0.161\linewidth]{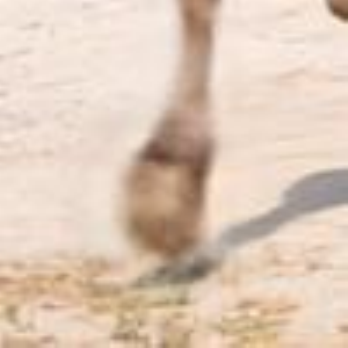}}\;\!\!
    	\subfloat[Ours]
    	{\includegraphics[width=0.161\linewidth]{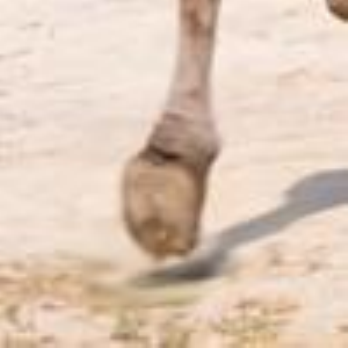}}\;\!\!
    	\subfloat[GT]
		{\includegraphics[width=0.161\linewidth]{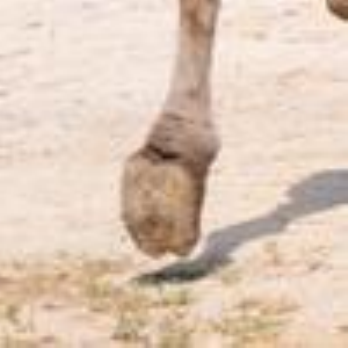}}
 \end{center}
\vspace{-2mm}
\caption{Examples interpolated by best-performing VFI methods.}
\label{fig:visual}
\vspace{-2mm}
\end{figure*}
%=====================================================================

\noindent\textbf{Training datasets.} We use the Vimeo-90k (septuplet) dataset~\cite{xue2019video} as well as the BVI-DVC dataset~\cite{ma2020bvi} to train our model. We adopt the same process of training data generation as in \cite{danier2021spatio}, obtaining more than 100K frame (256$\times$256) quintuplets in the form $(I_0,I_1,I^\mathrm{gt}_t,I_2,I_3)$ where $t=1.5$ (for $\times$2 frame interpolation).

% =================================================================
\begin{table}[t]
	\begin{center}
	\caption{Quantitative comparison results (PSNR/SSIM) for our model and 12 tested methods. OOM denotes cases where our GPU runs out of memory for the evaluation.}
\resizebox{\linewidth}{!}{
\begin{tabular}{lcccccc}
\toprule
     & \multicolumn{2}{c}{UCF101} & \multicolumn{2}{c}{DAVIS}& \multicolumn{2}{c}{VFITex}\\ \cmidrule(lr){2-3}  \cmidrule(lr){4-5}  \cmidrule(lr){6-7}
 & PSNR & SSIM & PSNR & SSIM & PSNR & SSIM \\ \midrule
				DVF~\cite{liu2017video} & 32.251 & 0.965  & 20.403 & 0.673  & 19.946 & 0.709 \\
				SuperSloMo~\cite{jiang2018super} & 32.547 & 0.968 & 26.523 & 0.866  & 27.914 & 0.911 \\
				SepConv~\cite{niklaus2017video} & 32.524 & 0.968 & 26.441 & 0.853 & 27.635 & 0.907 \\
				DAIN~\cite{bao2019depth} & 32.080 & 0.965 & 27.086 & \textcolor{blue}{0.873} & OOM & OOM \\
				BMBC~\cite{park2020bmbc} & 32.576 & 0.968 & 26.835 & 0.869 & OOM & OOM \\
				AdaCoF~\cite{lee2020adacof} & 32.488 & 0.968 & 26.445 & 0.854 & 27.639 & 0.904 \\
				FeFlow~\cite{gui2020featureflow} & 32.520 & 0.967 & 26.555 & 0.856 & OOM & OOM \\
				CDFI~\cite{ding2021cdfi} & 32.541 & 0.968 & 26.471 & 0.857 & 27.576 & 0.906 \\
				CAIN~\cite{choi2020channel} & 32.133 & 0.965 & 26.477 & 0.857 & 28.184 & 0.911 \\
				EDSC~\cite{cheng2021multiple} & 32.460 & 0.969 & 26.968 & 0.860 & 27.641 & 0.904 \\
				XVFI~\cite{sim2021xvfi} & 32.224 & 0.966 & 26.565 & 0.863 & 27.759 & 0.909 \\
				FLAVR~\cite{kalluri2020flavr} & \textcolor{blue}{33.242} & \textcolor{blue}{0.970} & \textcolor{blue}{27.450} & \textcolor{blue}{0.873} & \textcolor{blue}{28.487} & \textcolor{blue}{0.915} \\
				\midrule
				Ours & \textcolor{red}{33.431} & \textcolor{red}{0.970} & \textcolor{red}{27.633} & \textcolor{red}{0.878}  & \textcolor{red}{28.540} & \textcolor{red}{0.916} \\
				\bottomrule
		\end{tabular}
}
		\label{quantitative}
		\vspace{-5mm}
	\end{center}
\end{table}
% =================================================================

\noindent\textbf{Evaluation datasets and metrics.}
Because our model requires four frames to interpolate one, the evaluation datasets should provide $I_0,I_1,I_2,I_3$ for each output $I_t^{gt}$. Therefore, we evaluate on the UCF101~\cite{soomro2012ucf101}, DAVIS~\cite{perazzi2016benchmark} and VFITex~\cite{danier2021spatio} datasets. While UCF101 consists of low-resolution content of size $225\times225$, DAVIS contains 480p sequences with large motion content. The VFITex dataset includes HD videos involving various dynamic textures and covering both large and complex motion types. To evaluate the performance of different methods, the most commonly used metrics for VFI, PSNR and SSIM~\cite{wang2004image}, are used to assess video quality.

\subsection{Ablation Study}

\noindent\textbf{Spatiotemporal CNN.} The effectiveness of the 3D convolutions used in our proposed model is investigated by replacing them with 2D convolutions while keeping the overall architecture. The resulting variant is denoted as Ours-\textit{2D}. It is observed from Table~\ref{tab:ablation} that the overall performance deteriorates in this case, especially on content with large motion (DAVIS) and dynamic textures (VFITex). This implies that the proposed 3D CNN evidently facilitates capturing motion dynamics, hence enhancing VFI performance.

\noindent\textbf{Coarse-to-fine multi-flow estimation.} We analyse the impact of the coarse-to-fine structure by reducing the number of iterative levels for multi-flow estimation. Specifically, we set the levels $L_0$ at which the coarsest flow estimation is performed to $2,1$ and $0$ to compare with the original $L_0=3$\footnote{Because we train the model using 256$\times$256 patches, when $L_0>3$, the spatial resolution becomes smaller than the number of multi-flows $M$.}. Table~\ref{tab:ablation} shows that as $L_0$ increases, the performance on all datasets improves steadily, indicating the effectiveness of the coarse-to-fine multi-flow prediction.

\subsection{Comparison with State-of-the-art Methods}

We benchmarked our method against 12 popular VFI models including: DVF~\cite{liu2017video}, SuperSloMo~\cite{jiang2018super}, SepConv~\cite{niklaus2017video}, DAIN~\cite{bao2019depth}, and more recent state of the arts including BMBC~\cite{park2020bmbc}, AdaCoF~\cite{lee2020adacof}, FeFlow~\cite{gui2020featureflow}, CDFI~\cite{ding2021cdfi}, CAIN~\cite{choi2020channel}, EDSC~\cite{cheng2021multiple}, XVFI~\cite{sim2021xvfi}, and FLAVR~\cite{kalluri2020flavr}. For a fair comparison, we re-trained all the models with the same training dataset as ours until convergence. The evaluation results are summarised in Table~\ref{quantitative}, where the best and second best scores in each column are highlighted in red and blue respectively. It it noted that the proposed method outperforms all the benchmark VFI methods on all test sets, demonstrating the robustness and generalisation ability of our model. Fig.~\ref{fig:visual} shows frames interpolated by several best-performing methods on DAVIS, where our proposed model produced the best result.

More visual examples for the ablation and benchmark experiments as well as our code can be found on our project page: \url{https://danier97.github.io/EDC}.

\section{Conclusion}\label{sec:conclusion}
This paper presents a novel 3D CNN architecture that exploits a coarse-to-fine structure to estimate multi-flows at multiple scales for deformable convolution-based video frame interpolation. While 3D convolutions enhance the modelling of complex inter-frame motion using spatio-temporal features, the coarse-to-fine scheme further improves the ability of the DefConv-based warping method to capture large motions. The proposed designs have been validated through ablation studies and comprehensive benchmarking experiments, which clearly indicate that the proposed model performs favourably against many state-of-the-art models.

%%%%%%%%% REFERENCES
{\small
\bibliographystyle{IEEEtran}
\bibliography{egbib}
}

\end{document}